%% file: root.tex
\documentclass[letterpaper, 10 pt, conference]{ieeeconf}  % Comment this line out if you need a4paper
\IEEEoverridecommandlockouts                              % This command is only needed if 
\usepackage{graphics} 
\usepackage{epsfig} 
\usepackage{times} 
\usepackage{amsmath}
\usepackage{amssymb}  
\usepackage{amsfonts}      
\usepackage{amstext} 
\usepackage{mathtools}
\usepackage{bbm}
\usepackage{float}
\usepackage{subcaption}

\usepackage{hyperref}       

\usepackage{cite}
\usepackage[capitalize, nameinlink]{cleveref} 
\usepackage{booktabs}       
\usepackage{makecell}
\usepackage{siunitx}
\usepackage{algorithm}
\usepackage[noend]{algorithmic}
\newcommand{\SWITCH}[1]{\STATE \textbf{switch} (#1)}
\newcommand{\CASE}[1]{\STATE \textbf{case} #1\textbf{:} \begin{ALC@g}}
\newcommand{\ENDCASE}{\end{ALC@g}}

\newcommand{\DEFAULT}{\STATE \textbf{default:} \begin{ALC@g}}
\newcommand{\ENDDEFAULT}{\end{ALC@g}}
\newcommand{\DEFAULTLINE}[1]{\STATE \textbf{default:} }
\usepackage[nolist]{acronym} 
\usepackage{hyphenat}
\usepackage{xspace}
\newcommand{\eg}{e.g.\@\xspace}
\newcommand{\ie}{i.e.\@\xspace}

\DeclareMathOperator{\diag}{diag}

\title{\LARGE \bf
A Modular Aerial System Based on Homogeneous Quadrotors with Fault-Tolerant Control}
\author{Mengguang Li, Kai Cui and Heinz Koeppl% <-this % stops a space
\thanks{*This work has been co-funded by the State of Hesse and HOLM as part of the ``Innovations in Logistics and Mobility'' programme of the Hessian Ministry of Economics, Energy, Transport and Housing (HA project no.: 1010/21-12) and the LOEWE initiative (Hesse, Germany) within the emergenCITY center.}% <-this % stops a space
\thanks{The authors are with the Department of Electrical Engineering and Information Technology, Technische Universität Darmstadt, 64289 Darmstadt, Germany.
        {\tt\small \{mengguang.li, kai.cui, heinz.koeppl\}@tu-darmstadt.de}}%
}
\begin{document}
\maketitle
\thispagestyle{empty}
\pagestyle{empty}
\input{main.tex}
\addtolength{\textheight}{-9cm}
\bibliographystyle{IEEEtran}
\bibliography{references}
\end{document}

%% file: main.tex
\begin{abstract}
The standard quadrotor is one of the most popular and widely used aerial vehicle of recent decades, offering great maneuverability with mechanical simplicity. However, the under-actuation characteristic limits its applications, especially when it comes to generating desired wrench with six degrees of freedom (DOF). Therefore, existing work often compromises between mechanical complexity and the controllable DOF of the aerial system. To take advantage of the mechanical simplicity of a standard quadrotor, we propose a modular aerial system, IdentiQuad, that combines only homogeneous quadrotor-based modules. Each IdentiQuad can be operated alone like a standard quadrotor, but at the same time allows task-specific assembly, increasing the controllable DOF of the system. Each module is interchangeable within its assembly. We also propose a general controller for different configurations of assemblies, capable of tolerating rotor failures and balancing the energy consumption of each module. The functionality and robustness of the system and its controller are validated using physics-based simulations for different assembly configurations.
\end{abstract}
\section{INTRODUCTION} \label{sec1}
Inspired by biological systems such as ants, bees and other cooperative organisms that collaborate adeptly to accomplish diverse tasks, re-configurable modular robotic systems \cite{9341129, 9811583, 9561610} have emerged as an important innovation in various fields. In particular, within the domain of aerial vehicles, the concept of a modular aerial system \cite{10161064} has gained significant traction and prominence as a cutting-edge focus within robotics research.

Mechanical simplicity and hardware development have led to multi-rotors playing a major role not only in research but also in many commercial applications. A widely used multi-rotor is the quadrotor, which is well suited to many applications due to its agility and low maintenance costs, such as photography, non-contact inspection, etc. Common quadrotors have a co-linear arrangement of rotors, resulting in an under-actuated system with coupled position and orientation dynamics. When it comes to applications that require physical contact \cite{8299552, 9462539}, the standard quadrotor is highly limited in its ability to fulfil these tasks compared to fixed-base ground robots.
\begin{figure}[t]
  \centering
  \includegraphics[width=\linewidth]{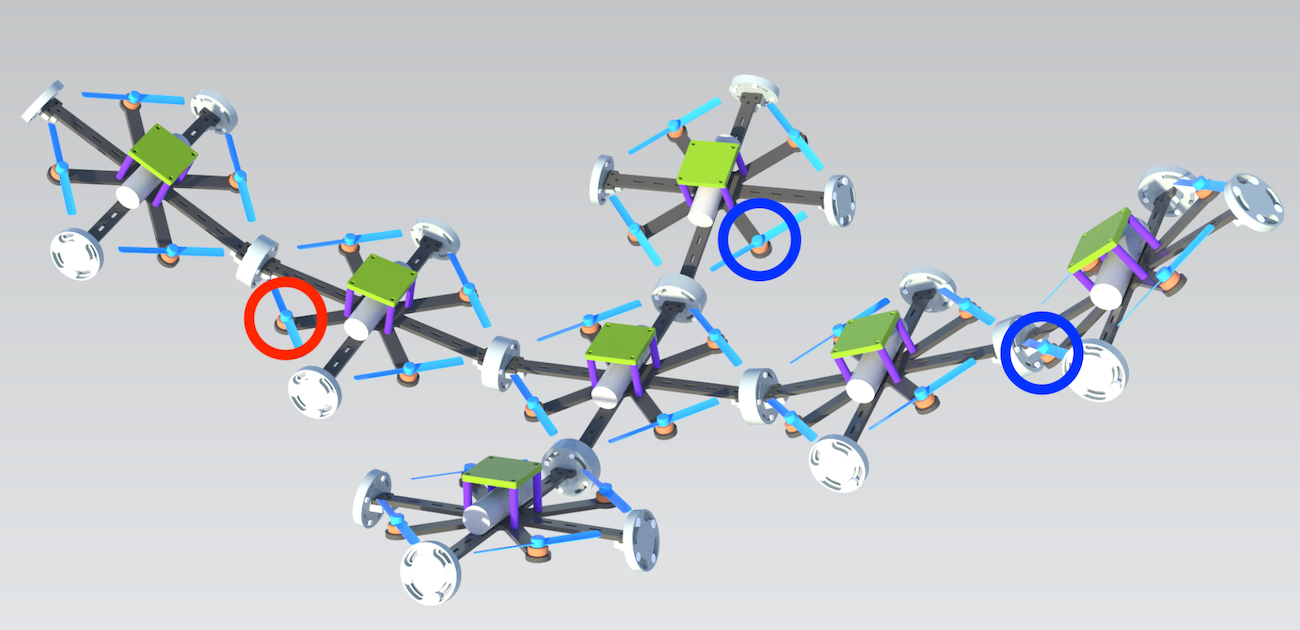}
  \caption{An assembly of $7$ IdentiQuads, where the rotor marked with a red circle fails at time $t=1s$ and the two rotors marked with dark blue circles fail at time $t = 5s$.}
  \label{fig: SIM 7 drones}
\end{figure}
Recent works have already focused on re-configurable designs with homogeneous or heterogeneous modular aerial robots. Saldaña et al. \cite{8461014} propose a modular aerial system based on quadrotors capable of assembling in mid-air, their updated version \cite{8758182} adding in-flight disassembly capability. A tetrahedral-shaped quadrotor design \cite{10161301} can be assembled vertically in a modular configuration. In \cite{5509882} a single rotor module is able to drive and dock with each other on the ground and fly. Although they can be configured in a modular fashion, the overall structures are still under-actuated, as all rotor orientations are co-linear, which does not change the controllable DOF of the system.

A straightforward way to achieve full actuation is to tile the rotor so that the orientation of the rotors is not co-linear, producing more controllable DOF. For non-modular systems, there are many existing multi-rotor designs, which can be divided into two categories, fixed tiling rotor \cite{7487497} and in-flight tiling rotor \cite{doi:10.1177/0278364920943654, 6868215}, we refer to \cite{doi:10.1177/02783649211025998} for a comprehensive review. Another possibility is to combine under-actuated multi-rotors into a new fully actuated structure. Pi et al. \cite{9658811} passively hinges four standard quadrotors to a six DOF platform, based on which Yu et al. \cite{9476994} place each quadrotor in a gimbal with two passive joints, for an over-actuated platform. Nguyen et al. \cite{8294249} also make use of standard quadrotors as thrust generators and passively connects them to a base with spherical joints, which is able to add more controllable DOF to the overall system. All these works focus more on full actuation and less on re-configurable modularity. The most related work to ours is the H-ModQuad \cite{9561016}, which proposes tilted quadrotors as re-configurable modules. These heterogeneous modules can form various structures to adapt to different types of tasks. However, the non-interchangeable modules and the non-standard mechanical design of the quadrotor add complexity to this approach. To focus on the homogeneity and simplicity of biological systems in nature, in this paper we instead use standard quadrotors as building blocks for our assembly.

The contributions of this work are threefold. First, we propose a modular aerial system based on homogeneous quadrotors capable of adding more controllable DOF up to six. Second, we propose a general controller that is suitable for all possible assembly configurations and is able to balance the energy consumption of each module to maximise the overall operating time. Third, we propose a fault-tolerant control strategy for the assembly in case of in-flight rotor failures. 
The paper is structured as follows. \cref{sec2} presents the design of IdentiQuad and derives the dynamics of the modular IdentiQuad and its assembly. \cref{sec3} determines the proposed general controller. \cref{sec4} evaluates the controller on different assemblies in physics-based simulations. \cref{sec5} concludes the work. 
\section{Design and Modeling} \label{sec2}
The focus of our work is to utilize standard quadrotors for a modular assembly, with less mechanical complexity while still allowing for more controllable DOF. Its design is detailed in the following. 
\subsection{Platform}
A single IdentiQuad module is based on a standard quadrotor with unidirectional rotors, \ie, the rotors only spin in one direction. An IdentiQuad is shown in \cref{fig:one module} with body-fixed frame $\{x, y, z\} $, we attach a cross-shaped frame underneath a standard quadrotor, each arm of the frame has the same length and is centered between two arms of the quadrotor. At the end of each attached frame, a replaceable connector is firmly attached to the frame. All the four connectors are identical. Each connector surface has an relative angle $\alpha \in [\pi/4, 3 \pi/4 ]$ to the $xy$-plane of the IdentiQuad frame. There are four magnets on each connector, each on its own arc of a circular track. Each magnet can be freely positioned and then fixed within its track for different configurations.
\subsection{Assembly}
\begin{figure}[t]
    \centering
    \hfill{}
    \includegraphics[height=3.2cm]{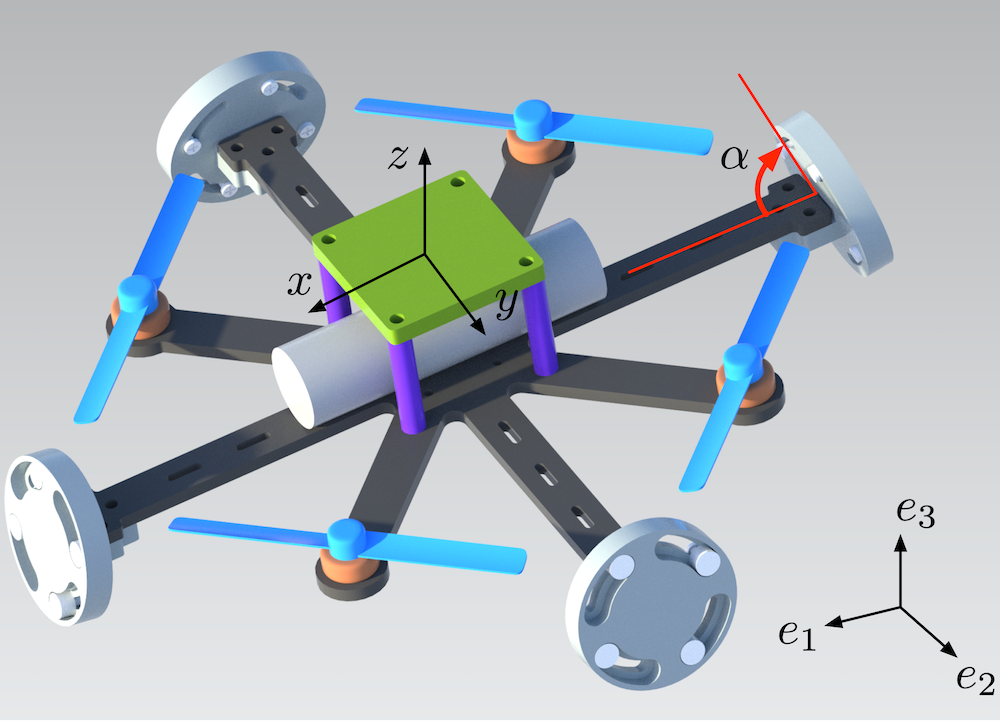}
    \hfill{}
    \includegraphics[height=3.2cm]{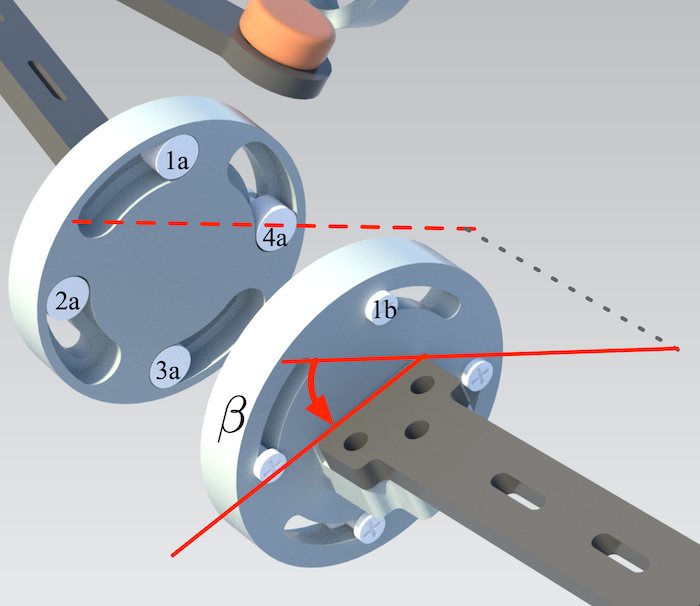}
    \hfill{}
    \caption{Design of an IdentiQuad module. Left: All four connectors have the same relative angle $\alpha$ to the frame. Right: A relative angle $\beta \in [0, 2 \pi)$ can be achieved by aligning magnet $1b$ with different magnets from $1a$ to $4a$.}
    \label{fig:one module}
\end{figure}

Each IdentiQuad can be connected to other IdentiQuad by one of its four connectors, and each pair of connectors can be rigidly connected by joining their rounded surfaces with four magnets. By rearranging the magnets on each connector, the mounting mechanism allows a relative angle $\beta \in [0, 2 \pi)$ between two connectors, as shown in \cref{fig:one module}. When $n$ IdentiQuads are assembled together, we call the structure an $n$-IdentiQuads assembly. Note that for $\alpha = \pi/2, \beta = 0$ the assembly reduces to ModQuad \cite{8461014}. With the two mounting angles, IdentiQuad allows for a wide range of assembly configurations. Despite the various assembly options, there are several limitations to be aware of for configuring an assembly. Firstly, as the system allows assembly in three-dimensional space, there may not be enough available space to mount a new IdentiQuad to an existing assembly, which needs to be considered when configuring a system. Secondly, the total thrust generated in the desired hovering direction should be at least greater than the total weight of the assembly. Thirdly, to minimise the downwash effect of the unidirectional rotor, no other IdentiQuad is to be mounted directly under the airflow of an IdentiQuad in an assembly. The configurations involving downwash will be investigated in future work.
\subsection{Dynamics}
As shown in \cref{fig:one module}, we denote the world frame as $\mathcal{F}_W$ with three unit basis vectors $\boldsymbol{e}_1 = \left[1, 0, 0 \right]^\top$, $\boldsymbol{e}_2 = \left[0, 1, 0 \right]^\top$, $\boldsymbol{e}_3 = \left[0, 0, 1 \right]^\top$. Since all the connections within the assembly are rigid, the resulting assembly can be considered a rigid body. We choose the center of mass (COM) of the assembly as the origin of the assembly frame. To better illustrate the controllable DOF of the assembly, we adapt the frame definition introduced by Xu et al. \cite{9561016}, which is an extension of the manipulability ellipsoid to the multi-rotor system. For a given assembly, the $z$-axis of the assembly frame is given in the direction of the maximum total thrust that can be generated. If the choice of $z$-axis is not unique, it suffices to arbitrarily select one of them as the $z$-axis direction. And the $x$-axis is in the direction of the maximum thrust that can be generated on the normal plane of the $z$-axis. Since we use a right-handed coordinate system in this paper, the $y$-axis can be easily determined. 

Given three axes, we could define the frame of the assembly as $\boldsymbol{R} \in SO(3)$. Each IdentiQuad $Q_i, i \in \{1, \dots, n \}$ within an assembly has four rotors $j \in \{1, \dots, 4 \}$, each rotor having an angular velocity of $\omega_{ij}$. The thrust and torque generated by a rotor can be expressed as $F_{ij}= c_F \omega_{ij}^2 $ and $M_{ij}= (-1)^j c_M \omega_{ij}^2 $, where $c_F$ and $c_M$ are rotor constants. The relative rotation from each quadrotor to the assembly is defined as $\boldsymbol{R}_i$, and each rotor's position relative to the assembly COM is defined as $\boldsymbol{o}_{ij}$. The direction of the total thrust generated by each quadrotor in the assembly frame can be expressed by $ \boldsymbol{q}_i = \boldsymbol{R}_i \boldsymbol{e}_3 $. The overall generated thrust and torque can be expressed as
\begin{align}
\begin{split}
\boldsymbol{T} &= \sum_{ij} F_{ij}  \boldsymbol{q}_i , \\
\boldsymbol{M} &= \sum_{ij} F_{ij} \boldsymbol{o}_{ij} \times \boldsymbol{q}_i + M_{ij} \boldsymbol{q}_i.
\end{split} \label{eq: F and M}
\end{align}
Since $ F_{ij}, M_{ij}$ both depend linearly on $\omega_{ij}^2$, we denote $\boldsymbol{u} = \left[\omega_{11}^2, \dots, \omega_{n4}^2 \right]^ \top$ as the control input for the assembly. The mapping from each rotor to the system can be written with the configuration vector $\boldsymbol{A}_{ij}$ according to \cref{eq: F and M},
\begin{equation}
\boldsymbol{A}_{ij} =  \left[ \begin{array}{c}
c_F \boldsymbol{q}_i  \\
c_F \boldsymbol{o}_{ij} \times \boldsymbol{q}_i  + (-1)^j c_M  \boldsymbol{q}_i \end{array} \right].
\end{equation}
Now \cref{eq: F and M} can be rewritten into a compact form
$\boldsymbol{A}_{n} \boldsymbol{u} = \left[ \boldsymbol{T}^\top , \boldsymbol{M}^\top \right] ^\top$, where $\boldsymbol{A}_{n} =  \left[ \boldsymbol{A}_{11}, \dots , \boldsymbol{A}_{n4} \right]$ is the assembly-dependent configuration matrix.
To further derive the dynamics, we denote the COM of the assembly as $\boldsymbol{p} = [p_x, p_y,p_z]^\top \in \mathbb R^3$, the angular velocity as $\boldsymbol \omega = [\omega_x, \omega_y, \omega_z]^\top$, the inertia tensor of each IdentiQuad as $\boldsymbol{J}_0$, the position of the $i$th module in the assembly as $(x_i,y_i,z_i)$ and the mass of one single IdentiQuad as $m$. Thus, the inertia matrix of an $n$-IdentiQuads assembly $\boldsymbol{J}$ can be calculated using the parallel axis theorem
\begin{displaymath}
    \boldsymbol{J} = \sum_i \mathbf R_i \boldsymbol{J}_0 \mathbf R^\top_i + m \left[ \begin{array}{ccc}
    y^2_i + z^2_i & -x_i y_i & -x_i z_i \\
    -y_i x_i  & x^2_i + z^2_i & -y_i z_i \\
    -z_i x_i & -z_i y_i  & x^2_i + y^2_i
    \end{array}  \right] .
\end{displaymath}
Therefore, the rigid body dynamics of the assembly can be described by Newton–Euler equations \cite{murray1994mathematical} as
\begin{equation}
    \begin{split}
        \ddot{\boldsymbol{p}} &= \frac{1}{n m} \boldsymbol{R} \boldsymbol{T} - g \boldsymbol{e}_3 \\
        \dot{\boldsymbol \omega} &= \boldsymbol{J}^{-1} (\boldsymbol{M} - \boldsymbol{\omega} \times \boldsymbol{J} \boldsymbol{\omega}) \\
        \dot{\boldsymbol{R}} &= \boldsymbol{R} [\omega]_\times ,
    \end{split}
% \end{align}
\end{equation}
where $[ \omega ]_\times$ defines the skew-symmetric matrix of the angular velocity $\omega$ as
\begin{displaymath}
    [\omega]_\times \coloneqq \left[ \begin{array}{ccc}
0 & -\omega_x & \omega_y \\
\omega_z & 0 & -\omega_x \\
-\omega_y & \omega_x & 0 \end{array} \right].
\end{displaymath}
\subsection{Objective}
An assembly can be configured with multiple modules to provide the payload or wrench required for a particular task. The first objective is to control the assembly to perform the task. As with re-configurable modules, we have the ability to add actuation redundancies to the assembly. It is then natural to design a robust system so that if one or more rotors fails, the system should still be able to perform the task. At the same time, the energy consumption of each module should be balanced so that the assembly can achieve maximum operating time. We address the above objectives in \cref{sec3}
\section{CONTROL} \label{sec3}
% first part: change \boldsymbol{R}_d, so that for all assembly to work
In this paper we focus on trajectory tracking control of the IdentiQuad assembly, which is required to follow a predefined trajectory of its COM position $\boldsymbol{p}$ and the assembly frame orientation $\boldsymbol{R}$. For a given desired trajectory $\left[ \boldsymbol{p}_d, \boldsymbol{R}_d \right]^\top \in SE (3)$, we apply a geometric controller \cite{5717652, 5980409} to control the IdentiQuad assembly. For computing the desired thrust, we apply a PD controller with feed-forward term
\begin{equation}
    \boldsymbol{T}_d = K_P(\boldsymbol{p}_d-\boldsymbol{p}) + K_D(\dot{\boldsymbol{p}}_d-\dot{\boldsymbol{p}}) + n m \ddot{\boldsymbol{p}}_d + n mg \boldsymbol{e}_3,
    \label{eq: T_d}
\end{equation}
where $K_P, K_D$ are diagonal matrices with positive constants. We obtain the error of rotation matrix as $\boldsymbol{e}_R = \frac{1}{2} [\boldsymbol{R}_d^\top \boldsymbol{R} - \boldsymbol{R}^\top \boldsymbol{R}_d] ^ \vee$, where $[\cdot]^\vee$ is the inverse of $[\cdot]_\times$, mapping a skew-symmetric matrix to a vector in $\mathbb{R}^3$. The error in angular velocity can be calculated by $\boldsymbol{e}_{\omega} =\boldsymbol{\omega} -  \boldsymbol{R}^\top \boldsymbol{R}_d \boldsymbol{\omega}_d$, which leads to the desired torque for the assembly
\begin{equation}
    \boldsymbol{M}_d = - K_R \boldsymbol{e}_R - K_{\omega} \boldsymbol{e}_{\omega} + \boldsymbol{\omega} \times \boldsymbol{J} \boldsymbol{\omega},
    \label{eq: M_d}
\end{equation}
where $K_R, K_{\omega}$ are diagonal matrices with positive constants.
For any given trajectory in three-dimensional space, $[\boldsymbol{p}_d, \boldsymbol{R}_d]^{\top}$ has six DOF, which requires 6 controllable DOF for tracking. As the assemblies can be configured differently, their controllable DOF can vary between four and six. We choose to always track the desired position $\boldsymbol{p}_d(t)$, while the desired orientation $\boldsymbol{R}_d(t)$ is tracked depending on the controllable DOF of the assembly. 
\subsection{Computing Desired Orientation}
\subsubsection{Four controllable DOF} \label{subsec four DOF}
When all the rotors in an assembly are co-linear, \ie, the row rank of $\boldsymbol{A}_n$ is four \cite{doi:10.1177/02783649211025998}, the system has four controllable DOF, \eg, a single IdentiQuad or multiple IdentiQuads with $\alpha = \pi/2, \beta = k \pi, k \in \{0, 1\}$. We choose to consistently track the desired position $\boldsymbol{p}_d$ and yaw angle $\psi_{d}$. With a given desired translational thrust $\boldsymbol{T}_d$, the $z$-axis of the desired orientation is determined by $\boldsymbol{z}_d = \boldsymbol{T}_d / \| \boldsymbol{T}_d \|$. Given the desired $\psi_d$, the desired $x$-axis direction can be obtained by $ \boldsymbol{x}_d = \boldsymbol{R}_{\boldsymbol{z}} (\psi_d) \boldsymbol{e}_1$, where $\boldsymbol{R}_{\boldsymbol{z}}(\psi_d)$ denote a basic rotation by an angle $\psi_d$ about the $z$-axis. The desired orientation can be computed by $\boldsymbol{R}_d = [\boldsymbol{y}_d \times \boldsymbol{z}_d, \boldsymbol{y}_d, \boldsymbol{z}_d]$, where $\boldsymbol{y}_d = (\boldsymbol{z}_d \times \boldsymbol{x}_d)/\| \boldsymbol{z}_d \times \boldsymbol{x}_d \|$.
\subsubsection{Five controllable DOF} \label{subsec five DOF}
For assemblies with a configuration matrix $\boldsymbol{A}_n$ row rank of five, there is one more DOF to track besides the desired position and the yaw angle. The desired pitch angle $\theta_d$ is selected for tracking, since the assembly can independently generate the desired thrust on the $x$-axis of the assembly frame. As with the case of four controllable DOF assembly, the roll angle is coupled with the position control for tracking the desired position $\boldsymbol{p}_d$. We would like to find out the desired $z$-axis orientation for obtaining the desired orientation $\boldsymbol{R}_d$. The coupling indicating that, for a given desired thrust $\boldsymbol{T}_d = [T_{x,d}, T_{y,d}, T_{z, d}]^{\top}$, the desired $z$-axis orientation $\boldsymbol{z}_d$ should be within the plane spanned by the two vectors $\boldsymbol{e}_1$ and $\hat{\boldsymbol{n}} = [0, T_{y,d}, T_{z,d}]^{\top}$. With the desired pitch angle $\theta_d$, we could rotate $\hat{\boldsymbol{n}}$ along the vector $\boldsymbol{m} = \hat{\boldsymbol{n}} \times \boldsymbol{e}_1 $, to obtain the desired $z$-axis orientation $ \boldsymbol{z}_d = \boldsymbol{R}_{\boldsymbol{m}}(\theta_d) \hat{\boldsymbol{n}} / \| \boldsymbol{R}_{\boldsymbol{m}}(\theta_d) \hat{\boldsymbol{n}}  \| $. As the $\boldsymbol{z}_d$ is defined, the desired orientation can be calculated using the same logic as above in \cref{subsec four DOF}.
\subsubsection{Six controllable DOF}
When $\boldsymbol{A}_n$ has full row rank, the assembly is a fully actuated system. The desired thrust and torque can be mapped independently to the rotor control inputs, without any coupling with the position control as in \cref{subsec four DOF,subsec five DOF}. Note that in this paper we consider trajectories that do not cause the rotor to saturate for assemblies without rotor failures. For assemblies with rotor saturation caused by rotor failures, we address that in \cref{fault tolerant}.
\subsection{Energy Balancing} \label{energy balancing}
With the desired orientation computed, the desired thrust and torque applied on the assembly can be obtained by \cref{eq: T_d,eq: M_d}. The control problem can be formalized as an allocation problem by solving the rotor control inputs $\boldsymbol{A}_{n} \boldsymbol{u} = \left[ \boldsymbol{T}_d^\top, \boldsymbol{M}_d^\top \right]^\top$ to track the desired wrench. The approach in \cite{9561016} applies the Moore–Penrose inverse to the configuration matrix $\boldsymbol{A}_n$ to minimize the sum of squares of the control inputs, \ie, the instantaneous consumption of the system. However, in a modular system, it is essential to balance energy consumption between the modules, as the batteries of each module are not shared by the system. For instance, if some rotors always spin faster than other rotors, or if a module starts with a lower battery voltage, this can cause the entire system to fail.

To solve this problem by optimization, we introduce a weighting function for four rotors of each $i$th module. By monitoring the battery voltage $V_i, i \in \{ 1, \dots, n \}$ of each IdentiQuad in an assembly, we denote
\begin{displaymath}
    H_{ij} = 1 + w \frac{ \Bar{V} - V_i}{\Bar{V}}, \quad \forall j \in \{1, \ldots, 4\} 
\end{displaymath}
where $w > 0 $ is a scaling constant and $\Bar{V}$ is the average battery voltage of the assembly. Modules with a lower weight are allowed to rotate faster, while a higher weight indicates a lower battery voltage, such that the corresponding rotors should rotate slower to consume less energy than the average. For all four rotors on the same module, we assign the same weight. To minimize the energy consumption while keeping a balanced battery voltage among modules, the control problem is formalized as a Tikhonov regularized optimization problem
\begin{displaymath}
    \min \Vert \boldsymbol{A}_{n} \boldsymbol{u} - \boldsymbol{b} \Vert^2_2 + \delta \Vert \boldsymbol{H} \boldsymbol{u} \Vert^2_2
\end{displaymath}
where
$\boldsymbol{H} = \diag(H_{11}, \dots, H_{n4}) \in \mathbb{R}^{4n \times 4n}$, $ \boldsymbol{b} = \left[ \boldsymbol{T}_d^\top, \boldsymbol{M}_d^\top \right]^\top$ and $\delta > 0$ is the regulation constant. The optimal control inputs can be obtained straightforwardly from the analytical solution
\begin{equation}
  \boldsymbol{u}^\ast = \boldsymbol{H}^{-2} \boldsymbol{A}_n^\top (\boldsymbol{A}_n \boldsymbol{H}^{-2} \boldsymbol{A}_n^\top + \delta \boldsymbol{I} )^{-1} \left[ \boldsymbol{T}_d^\top, \boldsymbol{M}_d^\top \right]^\top,  
  \label{eq: u*}
\end{equation}
where $\boldsymbol{I}$ is the identity matrix. 
\begin{figure}[t!]
    \centering
    \includegraphics[width=\linewidth]{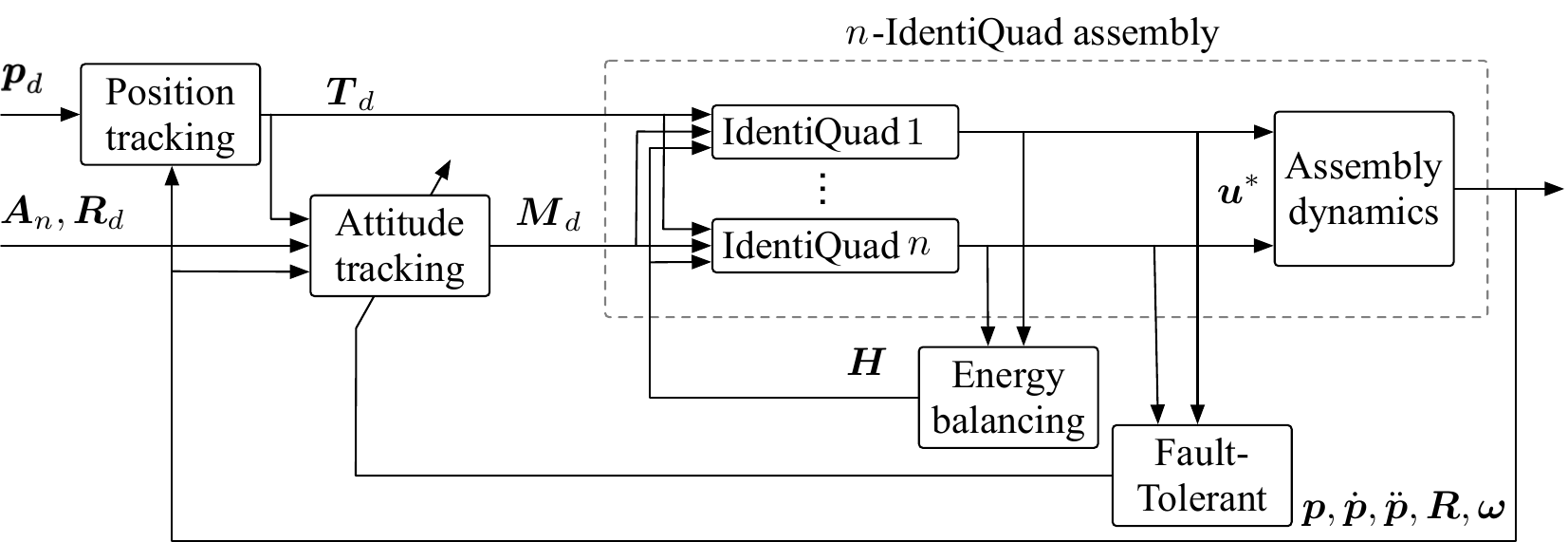}
    \caption{An overview of the proposed general controller for $n$-IdentiQuads assembly with energy balancing among modules and fault-tolerant in case of in-flight rotor failures.}
    \label{fig:control diagram}
\end{figure}
\subsection{Fault Tolerance} \label{fault tolerant}
For a modular system with adjustable actuation redundancy, we are inherently interested in a fault-tolerant control strategy in case of one or more rotor failures during operation. Unlike systems that adapt their configuration to rotor failures \cite{8977388}, we focus on in-flight rotor failures without changing the configuration of the assembly. We assume that the rotor failure is detected instantaneously and the controller reacts with a constant delay time of $\SI{2}{\milli\second}$, as in \cite{10179011}. The optimization formulation in \cref{energy balancing} provides a direct, weighted mapping from control inputs to desired wrench, which can easily be modified by removing entries of control inputs. For instance, when detecting the $ij$th rotor failure, we simply remove the $(j + 4i - 4)$th row and column from the weight matrix $\boldsymbol{H}$, $u_{ij} $ from the control input vector, and $\boldsymbol{A}_{ij}$ from the configuration matrix $\boldsymbol{A}_n$. The remaining terms then follow \cref{eq: u*} to obtain the optimal control input with only the functioning rotors. However, this approach is only applicable if the remaining rotors have sufficient redundancy so that they are still able to produce the desired wrench to track the given trajectory. For some assemblies, after rotor failures, reaching the same desired wrench could easily exceed rotor saturation, resulting in system failure. We propose an additional fault-tolerant control strategy to address this issue. 

For assemblies with six controllable DOFs, after a rotor failure is detected, we remove the entries related to the failed rotor as described in the last paragraph, and solve \cref{eq: u*}. As long as one of the control input $u_{ij}^\ast$ is reaching the rotor saturation upper-bound $u_{max}$, we propose to forego the desired row angle $\phi_d$. Instead we track only the remaining five DOF and compute the new desired orientation $\boldsymbol{R}_d$ as in \cref{subsec five DOF}. After solving \cref{eq: u*} we directly examine whether the control input still reaches rotor saturation. If so, we continue to disregard the desired pitch angle $\theta_d$ and track only the desired position $\boldsymbol{p}_d$ and the yaw angle $\psi_d$. This also applies to assemblies with five controllable DOF. In this fault-tolerant control strategy, position tracking is given a higher priority than orientation tracking, which is essential for a floating-base aerial system \cite{9720967}. A functioning position tracking in three-dimensional space provides the possibility to plan an emergency landing after rotor failures. The algorithm is described in \cref{algo}. Note that in this work, we only consider assemblies that can still follow a trajectory with at least four DOF after a rotor failure. Assemblies with less than 4 controllable DOF after rotor failures are left to future work. 
\begin{algorithm}[b!]
    \caption{Fault-Tolerant Control} \label{algo} 
    \begin{algorithmic}[1]
        \STATE Input: $\{ \boldsymbol{A}_n, \boldsymbol{H}, \boldsymbol{R}, \boldsymbol{T}_d, \boldsymbol{R}_d \}$.
        \IF {detected failure of the rotor $u_{ij}$}
            \STATE $\boldsymbol{A}_n \gets \boldsymbol{A}_n $ remove $\boldsymbol{A}_{ij}$; 
            \STATE $\boldsymbol{H} \gets \boldsymbol{H}$ remove $(j + 4i - 4)$th row and column;
            \STATE Controllable DOF $\gets rank (\boldsymbol{A}_n)$
            \ENDIF
            \SWITCH {Controllable DOF}
            \CASE {6}
              \STATE Compute $\boldsymbol{M}_d$ using \cref{eq: M_d}, $\boldsymbol{u}^{\ast}$ using \cref{eq: u*};
              \IF{$ \max \boldsymbol{u}^{\ast} \geq u_{max} $}
               \STATE Controllable DOF $\gets$ Controllable DOF $ - 1$
                \STATE \textbf{goto case} 5; \textbf{break};
              \ENDIF
            \ENDCASE
            \CASE {5}
              \STATE Compute $\boldsymbol{R}_d$ as in \cref{subsec five DOF}, $\boldsymbol{M}_d$ using \cref{eq: M_d}, $\boldsymbol{u}^{\ast}$ using \cref{eq: u*};
              \IF{$ \max \boldsymbol{u}^{\ast} \geq u_{max} $}
              \STATE Controllable DOF $\gets$ Controllable DOF $ - 1$
                \STATE \textbf{goto default}; \textbf{break};
              \ENDIF
            \ENDCASE
            \DEFAULT
              \STATE Compute $\boldsymbol{R}_d$ as in \cref{subsec four DOF}, $\boldsymbol{M}_d$ using \cref{eq: M_d}, $\boldsymbol{u}^{\ast}$ using \cref{eq: u*}; \textbf{break};
            \ENDDEFAULT
            % \ENDSWITCH
        \RETURN Fault-Tolerant Control Input $\boldsymbol{u}^{\ast}$.
    \end{algorithmic}
\end{algorithm}
An overview of the control strategy is shown in \cref{fig:control diagram}. 
\begin{figure*}[t]
  \centering
  \hfill{}
  \includegraphics[height=4.2cm]{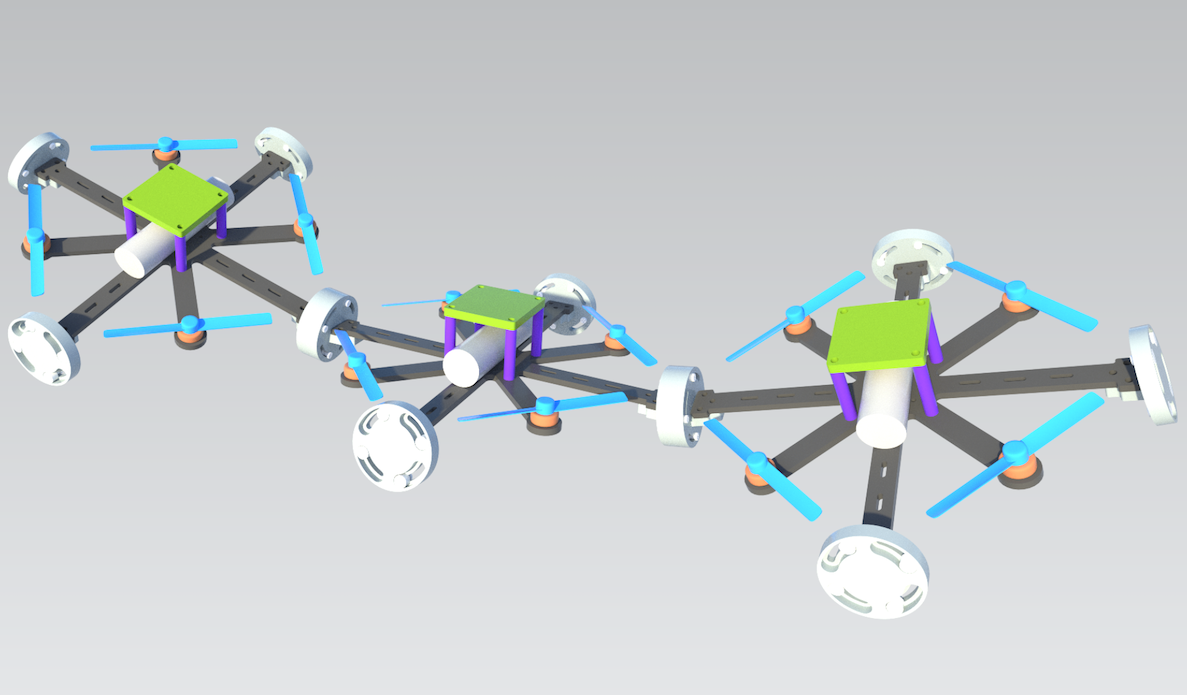}
  \hfill{}
  \includegraphics[height=4.2cm]{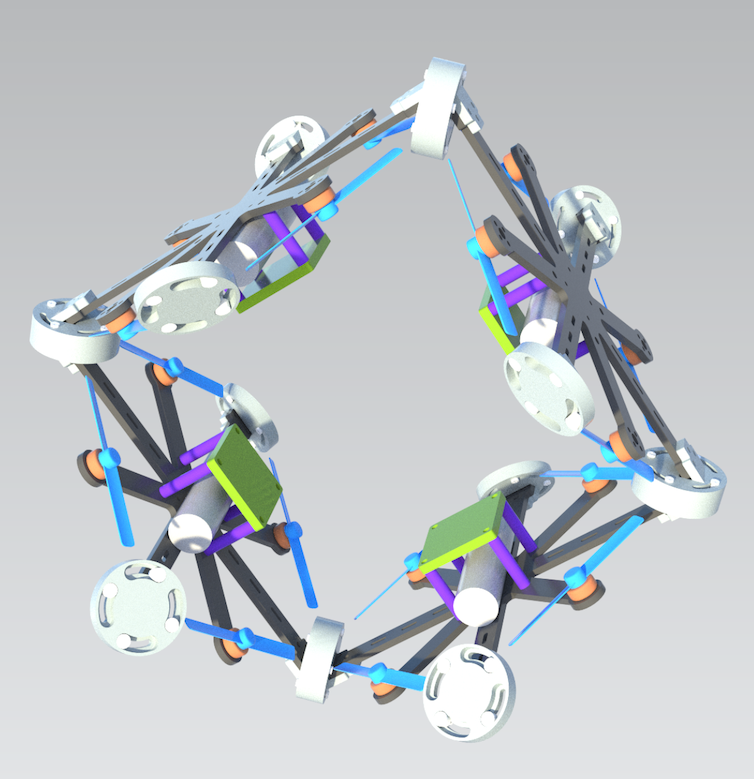}
  \hfill{}
  \includegraphics[height=4.2cm]{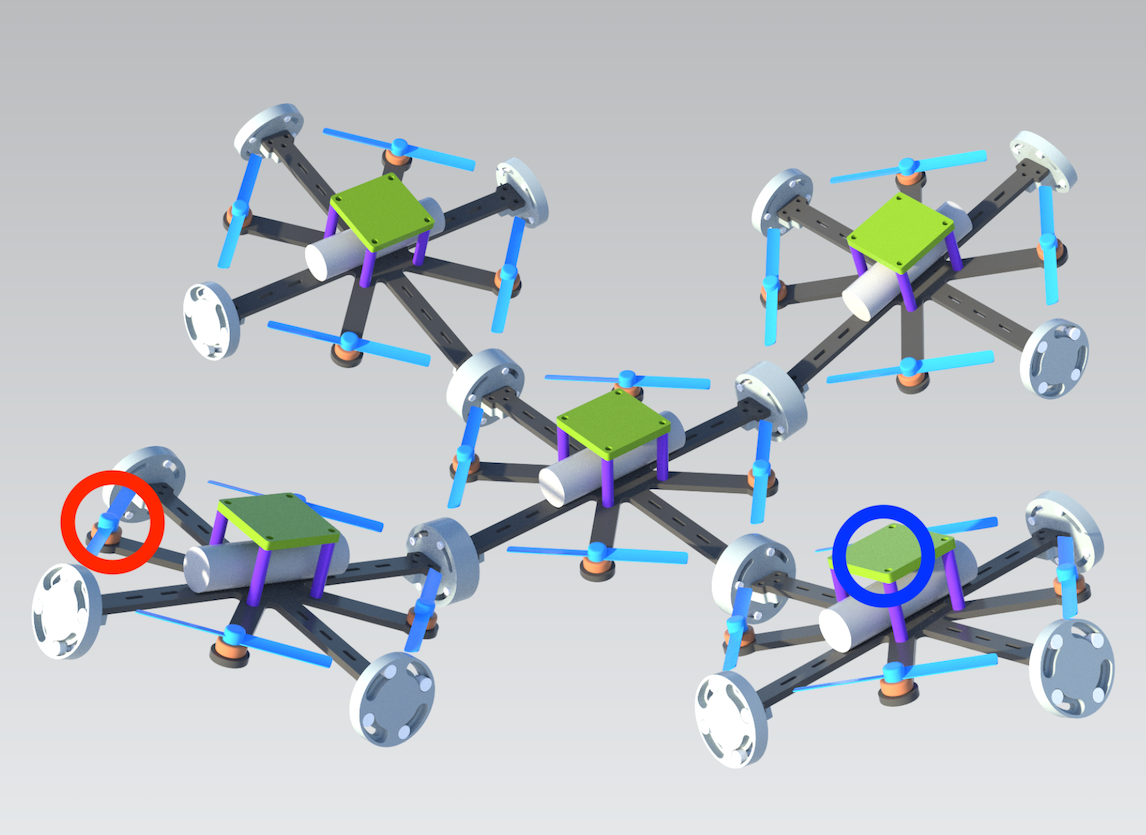}
  \caption{Left: A $3$-IdentiQuads assembly tracking three-dimensional figure-eight trajectory while maintaining its orientation. Middle: A $4$-IdentiQuads assembly tracking a five DOF trajectory with the desired roll angle $\phi_d$ range from $0$ to $2 \pi$, achieving a full circle angle tracking. Right: A $5$-IdentiQuads assembly start with different battery voltage, the dark blue marked module start with $70\%$ of its full battery voltage. The red circle marked rotor fails at time $t=5s$.}
  \label{fig:SIM 3 drones}
\end{figure*}
\section{EXPERIMENTS AND RESULTS} \label{sec4}
We experimentally evaluate the controller on different assemblies with varying tasks to demonstrate the functionality of the proposed work. The simulation is based on \texttt{\small gym-pybullet-drones} \cite{9635857}, a realistic multi-quadrotor simulator with great extensibility based on the Bullet physics engine.
\subsection{Six DOF trajectory tracking} \label{six DOF tracking}
\begin{figure}[t]
  \centering
  \includegraphics[width=.95\linewidth]{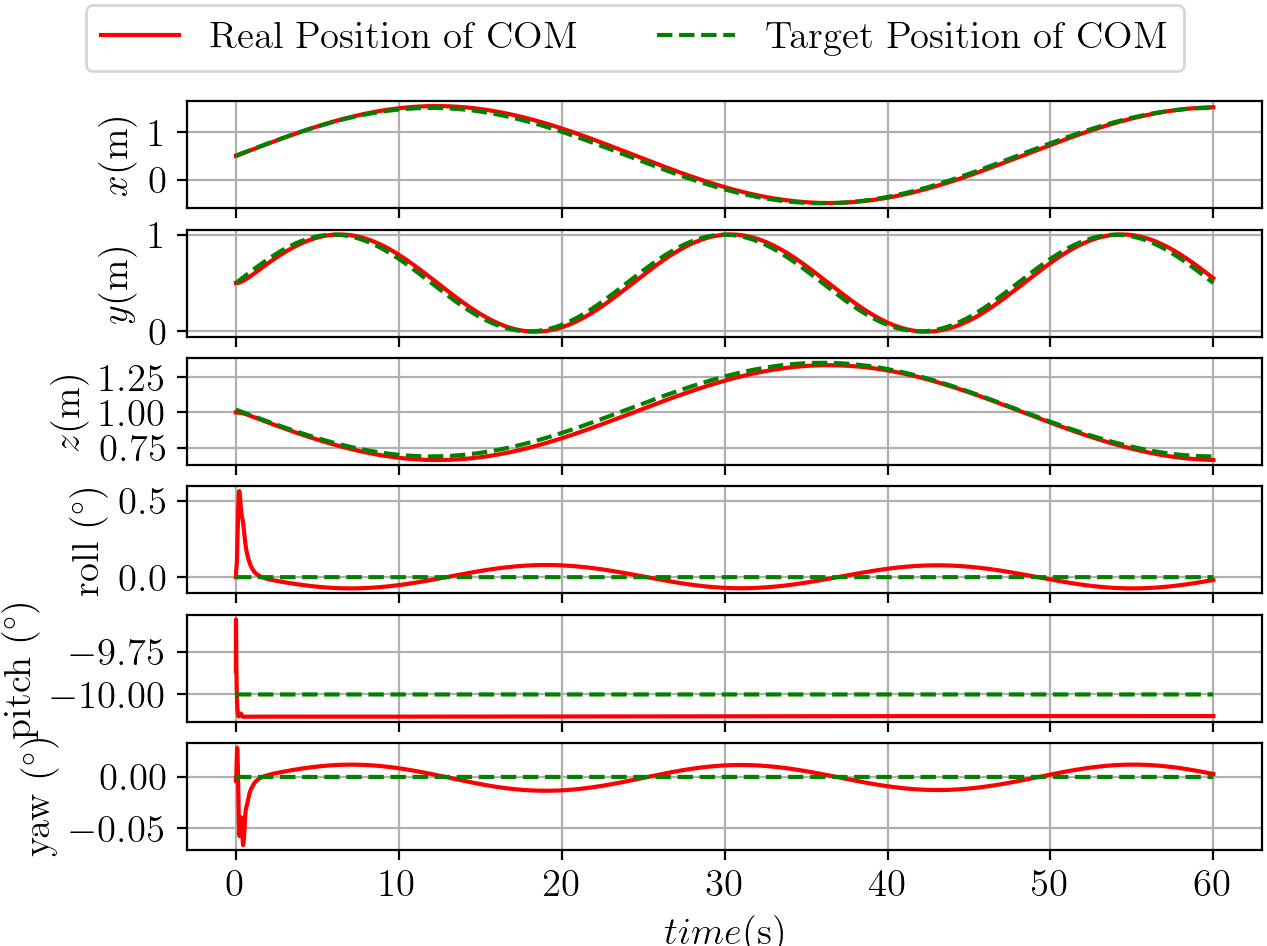}
  \caption{Simulation results of the $3$-IdentiQuads assembly tracking a six DOF trajectory.}
  \label{fig:SIM 3 drones results}
\end{figure}

To demonstrate the full actuation capability of the IdentiQuad assembly. With the two assembly angles $\alpha$ and $\beta$, we could use only three modules to achieve six controllable DOF. As shown in \cref{fig:SIM 3 drones}, three IdentiQuads with mounting angle $\alpha= 4 \pi/9 $ and a relative angle $\beta = \pi/9$ between each two modules. We assign a three-dimensional figure-eight trajectory as tracking task for the assembly $ f(t) = [l \sin(2\pi t), l \sin(2\pi t) \cos(2\pi t), - l/3 \sin(2 \pi t) ]$, where in this experiment $l = 0.2 m$. The result \cref{fig:SIM 3 drones results} shows that the assembly of only three IdentiQuads is able to track six DOF trajectory with relative small tracking error. 
\subsection{Horizontal full rotation tracking}
\begin{figure}[t]
  \centering
  \includegraphics[width=.95\linewidth]{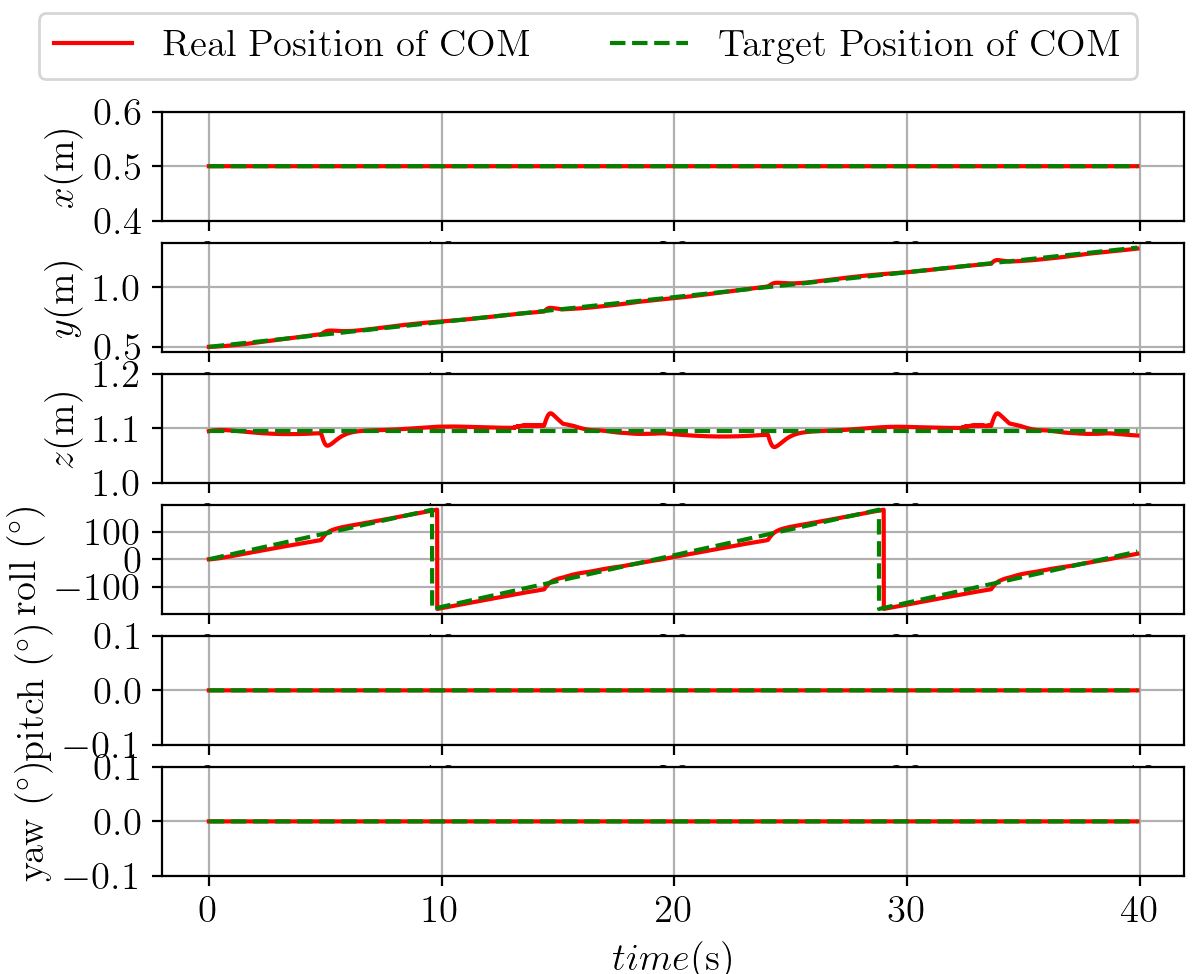}
  \caption{Simulation results of the $4$-IdentiQuads assembly tracking five DOF trajectory with desired roll angle $\phi_d$ rotating a full $2\pi$ rotation.}
  \label{fig:SIM 4 drones results}
\end{figure}

In this experiment, we use four modules with $\alpha= \pi/4, \beta = 0$ assembly as shown in \cref{fig:SIM 3 drones}, to track a five DOF trajectory, where the desired roll angle $\phi_d$ range from $0$ to $2 \pi$, achieving the tracking of a trajectory with a full $2\pi$ rotation. The result \cref{fig:SIM 4 drones results} shows that the assembly is able to track a full circle rotation horizontally. Note that the tracking performance is lower each time a module rotates near $\pi/2$, as the total thrust for the assembly in this phase comes mainly from one module, which suffers from motor saturation.
\subsection{Trajectory tracking with energy balancing}
\begin{figure}[b!]
  \centering
  \includegraphics[width=\linewidth]{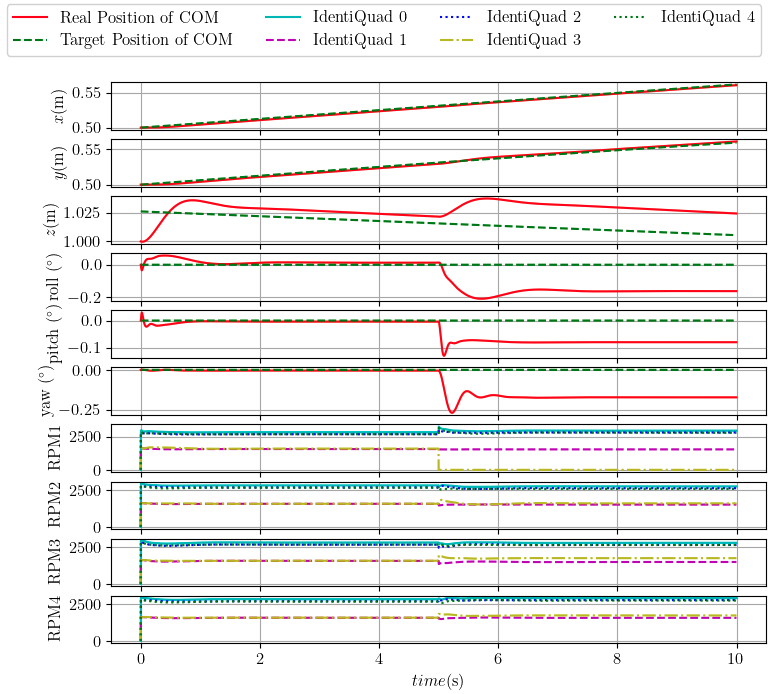}
  \caption{Simulation results of the $5$-IdentiQuads assembly start with different battery stage with rotor failure at time $t = 5s$.}
  \label{fig: SIM 5drones results}
\end{figure}

In this simulation task we evaluate the functionality of the energy balancing among modules. We assemble five IdentiQuads with a relative angle $\alpha = 2 \pi/5$, all the mounting angle $\beta$ are chosen to be $\beta = 0$. The tracking trajectory is the same as in \cref{six DOF tracking}. We chose one drone in \cref{fig:SIM 3 drones} marked with dark blue circle to start with $70 \%$ battery voltage, while the remaining four IdentiQuads start with the same $100 \%$ initial battery voltage. To verify the effectiveness of both energy balancing and fault-tolerant, we simulate one rotor failure at time $\SI{5}{\second}$. We choose $\SI{2}{\milli\second}$ as the communication delay time before the fault-tolerant control is called. As shown in \cref{fig: SIM 5drones results}, four rotors of the module with $70 \%$ battery voltage is rotating the slowest for a better energy balancing among the modules, while the assembly is still able to track the given six DOF trajectory. At time $t=\SI{5}{\second}$, the assembly is able to follow the trajectory despite a rotor failure under energy balancing.
\subsection{Trajectory tracking with rotor faults}
\begin{figure}[t]
  \centering
  \includegraphics[width=\linewidth]{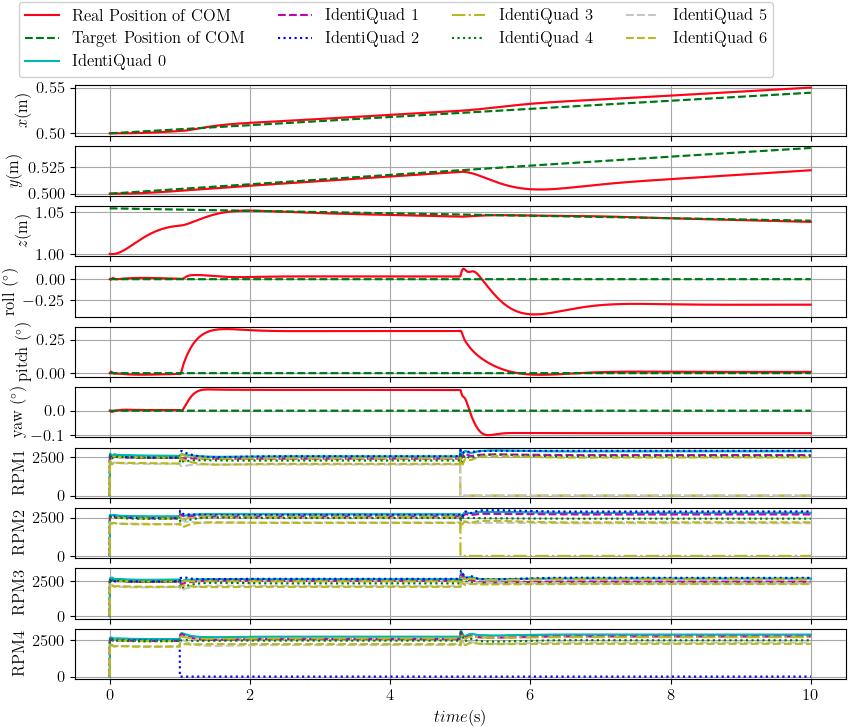}
  \caption{Simulation results of the $7$-IdentiQuads assembly with two different timing rotor failures, each at time $t = 1s$ and $t = 5s$.}
  \label{fig: SIM 7drones results}
\end{figure}

Lastly, we demonstrate the fault-tolerant control as described in \cref{algo}. We choose a $7$-IdentiQuads assembly with $\alpha = 4 \pi/9, \beta = 0$ as shown in \cref{fig:SIM 3 drones}. The tracking trajectory is the same as in \cref{six DOF tracking}. All the modules are assumed to start with the same battery voltage. At time $t = \SI{1}{\second}$, the first one rotor marked in red circle fails, and $\SI{4}{\second}$ later, two additional rotors marked in blue circle fail. The result is shown in \cref{fig: SIM 7drones results}. As the first rotor fails, the assembly is still able to generate six controllable DOF. While the yaw angle and $z$ position are affected by the fault, the assembly is able to track the original six DOF trajectory with a higher tracking error after stabilization. At time $\SI{5}{\second}$, as proposed in \cref{algo}, the assembly drops its tracking DOF twice, to maintain a steady position tracking while foregoing the desired roll and pitch angle. Although the remaining assembly with three rotor failures can still generate six DOF, the assembly would fail to keep tracking the original trajectory due to motor saturation. This experiment illustrates the effectiveness of the proposed fault-tolerant control strategy.
\section{CONCLUSIONS AND FUTURE WORKS} \label{sec5}
In this work, we propose an aerial assembly based on homogeneous modules, which have the ability to generate more controllable DOF. The system is extendable to various applications and is task-specifically modifiable. A general controller is proposed for different configurations of assemblies, which also aims to balance the energy consumption among modules. The proposed fault-tolerant strategy is able to maintain some degree of controllability of the assembly during in-flight rotor failures. Future work includes conducting field experiments with real quadrotors to verify the mechanical design, magnitude of the achievable wrench and robustness of the controller in practice. Additionally, the fault-tolerant controller could be extended to include assemblies with rotor failures resulting in fewer than four controllable DOF.